\documentclass[10pt,twocolumn,letterpaper]{article}
\usepackage[pagenumbers]{cvpr} 

\definecolor{cvprblue}{rgb}{0.21,0.49,0.74}
\usepackage[pagebackref,breaklinks,colorlinks,allcolors=cvprblue]{hyperref}

\def\paperID{3} 
\def\confName{CVPR}
\def\confYear{2025}

\usepackage{booktabs}
\usepackage[normalem]{ulem}
\usepackage{multirow}
\usepackage{cuted}

\makeatletter
\newcommand\xleftrightarrow[2][]{%
  \ext@arrow 9999{\longleftrightarrowfill@}{#1}{#2}}
\newcommand\longleftrightarrowfill@{%
  \arrowfill@\leftarrow\relbar\rightarrow}
\makeatother

\title{Multi-Flow:\\Multi-View-Enriched Normalizing Flows for Industrial Anomaly Detection}

\author{Mathis Kruse\qquad Bodo Rosenhahn \\
Institute for Information Processing,\; L3S - Leibniz University Hannover\\
{\tt\small kruse@tnt.uni-hannover.de}\\
}

\begin{document}
\def\method{\emph{Multi-Flow}}
\maketitle
\begin{abstract}
With more well-performing anomaly detection methods proposed, many of the single-view tasks have been solved to a relatively good degree.
However, real-world production scenarios often involve complex industrial products, whose properties may not be fully captured by one single image. 
While normalizing flow based approaches already work well in single-camera scenarios, they currently do not make use of the priors in multi-view data.
We aim to bridge this gap by using these flow-based models as a strong foundation and propose \method, a novel multi-view anomaly detection method.
\method~ makes use of a novel multi-view architecture, whose exact likelihood estimation is enhanced by fusing information across different views.
For this, we propose a new cross-view message-passing scheme, letting information flow between neighboring views. 
We empirically validate it on the real-world multi-view data set Real-IAD~\cite{realiad} and reach a new state-of-the-art, surpassing current baselines in both image-wise and sample-wise anomaly detection tasks.
\end{abstract}    
\section{Introduction}\label{sec:intro}

During any manufacturing process, a myriad of possible production errors inevitably lead to faults and costly replacement of products.
The field of industrial anomaly detection aims to solve this problem, by detecting anomalies in these products.
When detected early on in the manufacturing chain, this may successfully cut down on material and manual inspection cost, making the entire process more efficient.
As defects are not definable a priori and usually only occur very rarely, anomaly detection focuses on semi-supervised training schemes, where algorithms have access to normal data only~\cite{dataset:mvtec}.
This normal and anomaly-free data is often more widely available and can be more easily collected during already-running process pipelines. 

\begin{figure}
    \centering
    \includegraphics[width=0.95\linewidth]{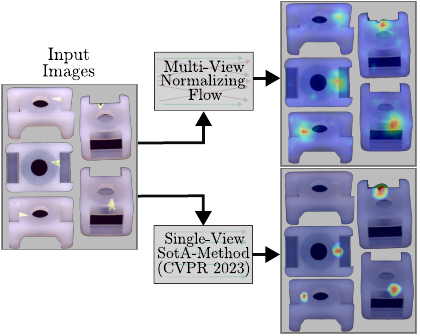}
    \caption{Overview of the multi-view enriched normalizing flow compared to SimpleNet~\cite{simplenet}. Multiple views of an object (with anomalies marked) get processed by the normalizing flow. Where single-view methods may struggle, \method~ detects anomalies irrespective of the objects view point.}
    \label{fig:teaser}
\end{figure}

Most image-based frameworks currently assume that only one static camera position is sufficient to determine whether an object is an anomaly~\cite{dataset:mvtec, dataset:mvtecloco, dataset:visa}.
Here, the most-used perspective is a birds-eye view of the object, as it is frequently used in data sets such as MVTec AD~\cite{dataset:mvtec}, MVTec Loco~\cite{dataset:mvtecloco} and the Visual Anomaly Dataset (VisA)~\cite{dataset:visa}.
This may neglect any three-dimensional or hidden information, which is only inferrable from other views, or by combining many views of one object.
The recently published data set Real-IAD~\cite{realiad} serves as an example of this new tasks.
Here, an object is now photographed from five different viewpoints, with one of them being the birds-eye view that is also present in other data sets.
While established methods may also work on images outside of the regular birds-eye camera pose, they fail to fuse the multi-view information to make an informed decision about the entire object~\cite{patchcore, simplenet}.

Existing multi-view methods, such as OmniAD~\cite{dataset:pad} or SplatPose~\cite{splatpose} use novel-view synthesis methods to model an object.
However, they require up to hundreds of views to capture the 3D information. Even then, they assume the existence of a perfect "normal" model, from which all anomalies deviate.
This is a crucial shortcoming, as it does not model any variance in texture or shape.

In this work, we propose \method, which manages to model a distribution across hundreds of different objects, each containing far fewer views (e.g. five for Real-IAD~\cite{realiad} compared to roughly two hundred for MAD~\cite{dataset:pad}).
This is achieved by fusing information across different views of an object using an architecture based on normalizing flows.
Normalizing flows have been shown to learn distributions of samples very well, which also makes them suitable to learn the anomaly-free normal distribution of our multi-view objects~\cite{glow_openai, csflow, differnet}.
For this, we specifically design a multi-view coupling block network that uses message-passing and aggregation of the multi-view data.
We further propose to leverage a data augmentation and conditioning scheme, to regularize and stabilize the training process.
Our method achieves state-of-the-art performance in the multi-view anomaly detection task on Real-IAD when detecting anomalies both at an image and at the instance level, while still achieving competitive results in segmenting the anomalies.
This makes us able to detect anomalies irrespective of the view they are in, as visualized in~\cref{fig:teaser}.
Here, SimpleNet~\cite{simplenet}, which does not model multi-view information, can be seen struggling with detecting all anomalies.
We also conduct several ablation studies to additionally motivate our choices of architecture and training procedure.

To summarize, our \textbf{contributions} include
\begin{itemize}
    \item We propose \method~for multi-view anomaly detection and reach a new state of the art for detection tasks on Real-IAD
    \item We design a multi-view-aware coupling block for efficient fusion of multi-view information
    \item We improve the training stability using a novel augmentation and conditioning scheme
    \item We make all code available at \url{https://github.com/m-kruse98/Multi-Flow}
\end{itemize}

\section{Related Work}\label{sec:related}

We review the field of anomaly detection and also give an overview of the research on normalizing flows.
\subsection{Anomaly detection}
Research on anomaly detection (AD) has differentiated into a number of varying types of algorithms.
Fundamentally, most algorithms are semi-supervised, as they only learn using normal, anomaly-free, data.

One strand of research uses \emph{reconstruction-based} methods.
These usually aim to reconstruct images, or features, through some information bottleneck, with the underlying assumption that anomalies will not be as easily reconstructed.
Early works used autoencoders~\cite{ae_ad, dae} or generative adversarial networks~\cite{ganomaly, anogan, defectgan, ocrgan}.
These models are especially prone to over-generalization, which may result in successful reconstruction for all types of inputs, contradicting the underlying assumptions of these methods.

Other methods have delved into \emph{incorporating anomalous samples} into the training process, sitting between semi-supervised and fully supervised approaches. They may use actual real-world anomaly samples~\cite{devnet} or self-generated and simulated anomalous data~\cite{draem, cutpaste, realnet, pni, simplenet} to fine-tune their decision boundaries. However, these methods need to be mindful of any biases they introduce towards their generated data.

Student-teacher networks have also seen a lot of usage in anomaly detection~\cite{rd, destseg, efficientAD, AST_rudolph}. In most of these, a student network is trained to mimic the behavior of a teacher network.
Since training is done on normal data, the student usually fails to mimic the teacher on anomalous data, thereby enabling the detection. Designing suitable tasks for the teacher is very important for these networks.

With the de-facto standard data set MVTec AD~\cite{dataset:mvtec} having reached near saturation in performance~\cite{patchcore, simplenet}, the research community has been looking for extensions to the current state-of-the-art.
Some authors try to fit the detection of all classes in the data sets into one single model, a task termed \emph{multi-class anomaly detection}~\cite{uniad, OmniAL, multiclass_vq}.
Other works focus on \emph{more difficult data sets}.
These may need algorithms to perform logical reasoning~\cite{dataset:mvtecloco} or work with other data modalities, such as 3D data, to detect anomalies~\cite{dataset:mvtec3d, dataset:real3d, dataset:anomaly_shapenet, dataset:eycandies}.
Other trends include detecting anomalies in \emph{multi-view data}.
Here, anomalies need to be detected in all views they appear in, irrespective of the object pose, making the task more difficult.
Pose-agnostic anomaly detection, as posed in the MAD data set~\cite{dataset:pad, dataset:RAD}, requires reasoning across dozens of views of one single object.
While embedding this data into the 3D space using novel-view synthesis methods has seen some success, conventional 2D methods still struggle heavily in this setting~\cite{splatpose, dataset:RAD}.
Another setting has been proposed with the Real-IAD data set~\cite{realiad}, which, inspired by real production scenarios and factory camera setups, features five fixed camera poses per object.
Contrary to MAD, which contains dozens of images of one specific normal instance~\cite{dataset:pad, dataset:RAD}, Real-IAD contains much sparser multi-view data of many different normal instances~\cite{realiad}.
Hence, Real-IAD requires reasoning across a huge range of multi-view instances, including their variation in texture and shape.

With the advent of large pre-trained foundation models for both image and text, a new branch surrounding few-shot or \emph{zero-shot anomaly detection} methods has formed~\cite{WINClip, AnomalyClip, AnomalyGpt}.
AnomalyGPT~\cite{AnomalyGpt} leverages existing feature-based methods and contextualizes and verbalizes the found anomalies using a Large Language Model (LLM). Works such as WinCLIP~\cite{WINClip} leverage the shared image-text latent space of CLIP embeddings~\cite{CLIP} to detect anomalous concepts without any training.
While there is progress on the standard detection tasks, there is still work to be done in adapting the zero-shot methods to detect anomalies in data modalities other than regular single-view images.

Lastly, \emph{density estimation} techniques have seen success in anomaly detection. These works leverage pre-trained features as their latent space on which they estimate a samples likelihood.
This may be done with parametric models such as a multivariate Gaussian~\cite{padim} or implicitly using nearest-neighbor methods~\cite{patchcore, cfa, SPADE}.
These methods are highly dependent on the choice of features, as they require distances in feature space to correspond to the difference between anomalies and normal samples~\cite{heckler_features}.

\subsection{Normalizing Flows} 
Normalizing Flows are a widely applied set of techniques for anomaly detection~\cite{differnet, csflow, cflow, fastflow, pyramidflow, vorausAD, nf_quantum}, falling into the category of density estimation.
These flows are invertible mappings between two distribution spaces.
During the forward direction, this enables the expression of likelihoods within a complex feature distribution by means of a simpler base distribution, such as a standard Gaussian~\cite{rezende_variational_nf}.
Due to the bijectivity of a flow, they can also be used in their backwards direction.
It is possible to generate a sample from the base distribution and use the reverse direction of the flow.
This generates a sample in the original complex distribution, giving the flow generative capabilities when training on e.g. images of human faces~\cite{glow_openai}. 
Nowadays, normalizing flows have fallen out of favor in many generative tasks, as diffusion models managed to achieve strong generation performances~\cite{ddpm_probabilistic}.

Still, normalizing flows fulfill an interesting niche when used for exact probability distribution modeling~\cite{cvpr24_nf_segmentation, differnet, pose_nf2} or for problems bijective in nature~\cite{pose_nf1}.
A lot of research has also gone into proposing efficient neural network architectures, while preserving invertibility, among other criteria, for normalizing flows.
Architectures using coupling blocks have seen wide application~\cite{NF_NICE, glow_openai}, with RealNVP~\cite{RealNVP} being used for its simplicity and efficient computation.
These methods partition the data and let these splits transform each other using learned (affine) transformations, all while preserving invertibility. Research has focused on proposing new expressive transformations such as ones based on rational quadratic splines~\cite{neural_spline_flows} or other nonlinear functions~\cite{flow_plusplus}.
Lastly, normalizing flows can also be used in an autoregressive fashion~\cite{nf:IAF, NF:MADE, NF:MAF}.
While these autoregressive methods enjoy strong performances in some tasks, they have higher computational demands than RealNVP-based architectures.
RealNVP only needs one forward pass for an estimate in either direction, whereas autoregressive models need to update their internal state, requiring several passes to compute the likelihood of a sample.
Depending on their architecture, either sampling~\cite{NF:MAF} or likelihood estimation~\cite{nf:IAF} become more costly than their RealNVP counterpart.

\section{Methodology}\label{sec:method}

We describe the full training and testing procedure for our approach.
It has been shown, that normalizing flows underperform in anomaly detection tasks, when training on  raw image data~\cite{nf_kirichenko_ood}.
We therefore learn our normalizing flow on a feature distribution $y \in \mathcal{Y}$, where we obtain each feature $y$ using a standard pre-trained feature extractor $f_\text{F}$ on input images $x \in \mathcal{X}$.
Our normalizing flow, parametrized using a neural network, then maps these features onto the Gaussian distribution $p_\mathcal{Z}$.
Therefore, we use the mappings
\begin{equation}
    x \in \mathcal{X} \xrightarrow[ f_F  ]{} y \in \mathcal{Y} \xleftrightarrow[f_{\theta}]{} z \in \mathcal{Z} \sim \mathcal{N}\left(0,I\right),
\end{equation}
where $f_F$ is the frozen feature extractor, and $f_\theta$ is our trainable bijective normalizing flow.
\subsection{Model Setup}

\begin{figure}
    \centering
    \includegraphics[width=\linewidth]{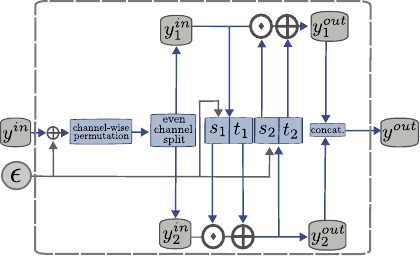}
    \caption{RealNVP~\cite{RealNVP} coupling block with conditioning. The input $y^{in}$ is augmented with a noise vector $\epsilon$ and split into two parts along its channel dimension. Each split and the conditioning noise component $\epsilon$ is concatenated and given to an $st$-network, which calculates components $s$ and $t$ for transforming the opposing path.}
    \label{fig:coupling_block}
\end{figure}

An image $x_i$ is passed through a pre-trained feature extractor to receive a feature vector $y_i \in \mathbb{R}^{C\times W\times H}$.
This vector serves as input to the flow model, which consists of several coupling blocks chained behind one another.
A diagram of the information flow in such a coupling block can be found in~\cref{fig:coupling_block}. 
It consists of permuting and splitting along the channel dimension, and calculating the affine transformation as
\begin{align}
    \begin{split}
        y_1^{out} &= y_1^{in} \odot e^{s_2\left(y_2^{out}\right)}  \oplus t_2\left(y_2^{out}\right), \;\text{and}\\ 
        y_2^{out} &= y_2^{in} \odot e^{s_1\left(y_1^{in}\right)}  \oplus t_1\left(y_1^{in}\right).
    \end{split}
\end{align}
Here, $\oplus$ and $\odot$ are element-wise addition and multiplication, respectively. The most integral part are the $st$-networks, which estimate additive and multiplicative element-wise components of the affine transformations.
They will be described in greater detail in~\cref{sec:st_network}.
To invert the entire model, additive components need to be subtracted, and multiplicative ones are divided.

We further apply a soft clamping to the scale coefficients $s$ to increase the training stability~\cite{ardizzone_nf} by constraining them to the interval of $\left[-\alpha, \alpha\right]$ using
\begin{equation}\label{eq:ardizzone}
    s_{\text{clamp}} = \frac{2\alpha}{\pi} \arctan\left( \frac{s}{\alpha}\right).
\end{equation}

Given a feature vector $y$ our model puts out a pixel-wise likelihood estimate $z$, which also serves as the models anomaly score.
Taking the maximum aggregates them into a single image-wise score.
Further, the maximum operator is also used when aggregating scores of several multi-view images into one sample-wise score.

\subsection{Background Removal}
\begin{figure}
    \centering
    \includegraphics[width=0.95\linewidth]{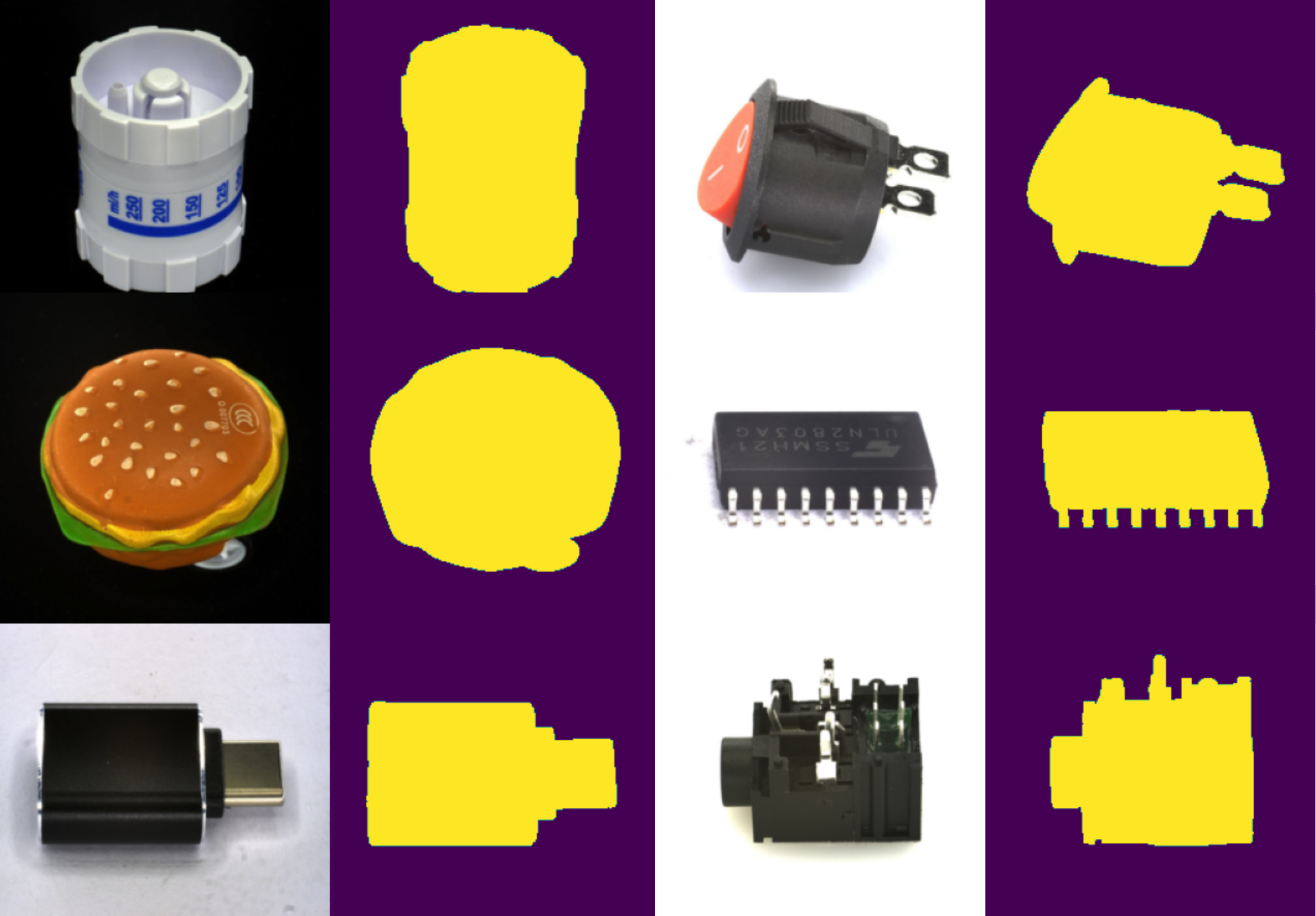}
    \caption{Examples of the background removal using the pre-trained dichotomous segmentation network MVANet~\cite{MVANet} on Real-IAD~\cite{realiad}. The usually homogeneous background regions can be extracted in all cases and without any grave mistakes.}
    \label{fig:background_removal}
\end{figure}

Currently, the model has to focus on the entire image feature space, including both foreground and background.
Any irregularities, such as background dust or other particles, may also be detected as an anomaly, despite not constituting an actual product-related defect.
Ignoring the background has been shown to improve detection results~\cite{AST_rudolph}.
Therefore, we employ the state-of-the-art dichotomous image-segmentation network MVANet~\cite{MVANet} to remove all background from foreground objects, as seen in~\cref{fig:background_removal}.
Here, we take the output of MVANet and apply a binary hole closing as well as an $8\times8$ dilation, to ensure getting one single-component to mask the entire object.
We resize the masks to the same size as the extracted feature maps for use in training.
By only calculating the loss on the foreground objects, we only estimate the object likelihood, ignoring the background completely.
Thereby, the network can fully focus on estimating the objects likelihood and will not be distracted by confounding factors in the background.

\subsection{Multi-View \texorpdfstring{$st$}{st}-Network Design}\label{sec:st_network}

\begin{figure*}
    \centering
    \includegraphics[width=0.8\textwidth]{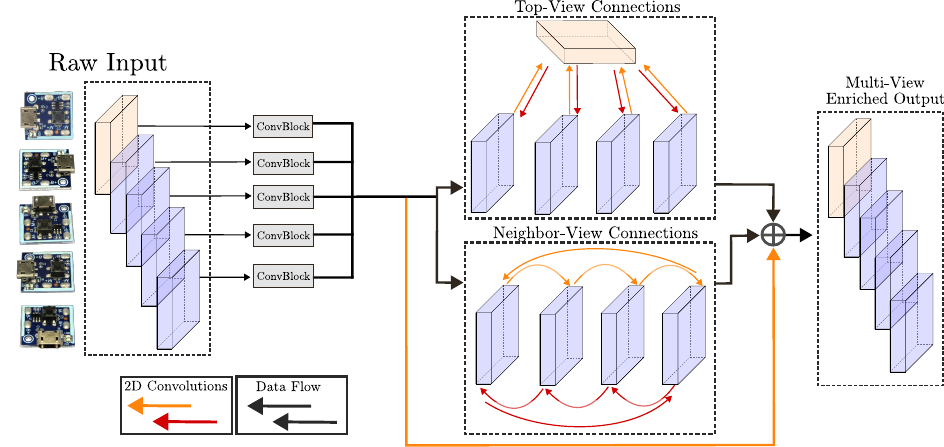}
    \caption{Architecture for sharing information across views. Each $st$-network implements one of these blocks. The raw input consists of a feature map for each of the object views. Each feature is finetuned by being passed through a ConvBlock. Then, subsequent 2D convolutions are applied and the data is aggregated according to the multi-view setup in~\cref{eq:neighbor_sums}. Top-view connections let information flow from the birds-eye view to all others. Neighbor-view connections lets the information flow between adjacent side views of the object.}
    \label{fig:crossview_architecture}
\end{figure*}

The way the invertible RealNVP coupling blocks are designed, as seen in \cref{fig:coupling_block}, networks $s$ and $t$ may be arbitrarily complex non-invertible neural networks.
Since all the computational power of the normalizing flow resides in the $st$-network, we need to carefully choose their design.

With our model focusing on detecting the anomalies in a sample-wise fashion, we group the input to our model into batches of images belonging to the same instance. 
The model receives batches of $n_v$ feature maps $\{y^i\}_{i=0}^{n_v}$ per instance, where each feature vector $y^i$ corresponds to the image of the $i$-th camera.
As for the case of Real-IAD, we therefore group together $n_v = 5$ different views.
This allows for reasoning across views, which regular anomaly detection methods do not take into consideration.
An overview of the paths of information flow is given in~\cref{fig:crossview_architecture}.

Each of the input vectors $y^i$ get passed through a convolutional block, which transforms them into a new representation $\hat{y}^i$. This simple convolutional block consists of a two-dimensional convolution with $5\times5$ kernels in the last coupling layer and $3\times3$ kernels in all preceding ones, as well as a standard ReLU nonlinearity.
The transformed vectors $\hat{y}^i$ then serve as input to all further connecting paths in the $st$-network.
Here, we aim to let information flow between neighboring views.
This enables anomalous features that are easy to see in one view, to influence others, where they may be much harder to find.
Still, we want to retain the models ability to also output separate anomaly maps for all input vectors, to enable the image-wise anomaly detection.

The most basic building block of all cross-view operations are 2D convolutions, denoted by $f_{\text{cross}}$.
Each representation $\hat{y}^i$ is enriched by information derived from all neighboring views.
Thus, each feature map gets infused with multi-view information by calculating
\begin{equation}\label{eq:neighbor_sums}
    y^i_{\text{output}} = f_\text{cross}^{ii}\left(\hat{y}^i\right) + \frac{1}{\left|N\left(\hat{y}^i\right)\right|} \sum_{j \in N\left(\hat{y}^i\right)} f_{\text{cross}}^{ji}\left( \hat{y}^j\right),
\end{equation}
where $N\left(\hat{y}^i\right)$ denotes the set of all neighboring views of $\hat{y}^i$ and $f_{\text{cross}}^{ji}$ is the convolutional layer connecting the $j$-th view to the $i$-th view.
We also consider every view $\hat{y}^i$ as connected to itself, to further stabilize the image-wise likelihood estimation.
While we define the exact camera neighbors for Real-IAD, this framework may be extended to other multi-view settings.

\paragraph{Top-View Connections.} In Real-IAD, all objects contain one birds-eye view, and four surrounding views~\cite{realiad}.
As visualized in~\cref{fig:crossview_architecture}, we consider the top-view a neighbor to all other surrounding views, letting information flow to all side-views and vice-versa.
In~\cref{fig:crossview_architecture}, each of the colored arrows here again denote a 2D convolution $f_\text{cross}$, which may represent the features according to the target-views needs.

\paragraph{Neighbor-View Connections.} Continuing with letting information flow between adjacent views, we also consider all surrounding views next to each other to be neighbors.
Again, each side-view passes and receives its information from the adjacent views on both the right and the left side.
While it may be possible that opposing views also see similar object contents, we do not force any connections between them.
Their relationships may instead be captured within the top-view connections of the architecture.

\subsection{Training Objective}

We want to maximize the likelihood of the training features $y \in \mathcal{Y}.$
This is done by transforming them into the Gaussian space $\mathcal{Z}$, where measurements of likelihood $p_{\mathcal{Z}}\left(z\right)$ are easily tractable.
When $f_\theta\left(y\right) = z$ is the output of our flow, we may express the feature distribution $p_\mathcal{Y}\left(y\right)$ using the \emph{change-of-variables formula} as
\begin{equation}\label{eq:change_of_variables}
    p_\mathcal{Y}\left(y\right) = p_\mathcal{Z}\left(z\right) \left|\det \frac{\partial z}{\partial y} \right|.
\end{equation}
Work such as SimpleNet~\cite{simplenet} add noise to the training samples, to better approximate the entire feature space, including anomalous features, while avoiding overfitting.
Following SoftFlow~\cite{softflow}, we sample noise $\epsilon$ during training and use them as additive components to perturb the training data points as
\begin{equation}
    x' = x + \epsilon, \;\text{where}\;\; \epsilon \sim \mathcal{N}\left(0, c^2 I\right).
\end{equation}
SimpleNet~\cite{simplenet} makes use of this noise to generate pseudo-anomalies to train a discriminator.
We do not explicitly model a discriminator but rather use this noise as a condition for the normalizing flow, letting it learn the conditional mapping $f_\theta\left(y | \epsilon\right) = z$.
This lets the flow more easily approximate the entire data manifold of both normal and anomalous samples~\cite{softflow}.
Since $p_\mathcal{Z}(z)$ is parametrized as a standard normal gaussian, maximum likelihood training using~\cref{eq:change_of_variables} simplifies to the loss function
\begin{equation}\label{eq:loss}
    \begin{split}
        \mathcal{L}\left(y\right) &= -\log p_\mathcal{Y}\left(y\right) = \frac{\left|\left|z\right|\right|_2^2}{2} - \left|\det \frac{\partial z}{\partial y} \right|, \\ 
        &\text{where}\;\; z =  f_\theta(y | \epsilon).
    \end{split}
\end{equation}
This is also the formula to calculate the anomaly maps with their respective anomaly scores, resulting in high values (i.e. low likelihood) for anomalous regions, and low values (i.e. high likelihoods) for normal regions.
With the architecture from RealNVP, the determinant of the Jacobian $\frac{\partial z}{\partial y}$ in~\cref{eq:loss} reduces to the sum of all scale values $s$ in the coupling blocks.

\section{Experiments}\label{sec:experiments}

\begin{table*}[ht]
\centering
\resizebox{\textwidth}{!}{
\begin{tabular}{l||c|c|c|c|c|c|c|c|c}
\multicolumn{1}{c||}{Metrics}    & PaDiM & Cflow & SoftPatch & DeSTSeg & RD   & UniAD & PatchCore    & SimpleNet &   \method          \\
\multicolumn{1}{c||}{AUROC ($\uparrow$)} & \cite{padim}  & \cite{cflow} & \cite{softpatch} & \cite{destseg} & \cite{rd} & \cite{uniad} & \cite{patchcore} & \cite{simplenet} & (ours) \\\hline\hline
Sample-wise & 91.2  & 89.8  & 92.8      & 94.0      & 83.7 & 88.1  & 93.7         & \uline{94.9}                & \textbf{95.85} $\pm$ 0.02   \\
Image-wise  & 80.3  & 82.5  & \uline{89.4}      & 86.9    & 87.1 & 82.9  & \uline{89.4} & 88.5            & \textbf{90.27} $\pm$ 0.02             \\
\end{tabular}
}
\caption{\textbf{Average} detection performance of various methods across all classes in Real-IAD~\cite{realiad}. Sample-wise denotes the setting, where the score of images of the same object are aggregated to one single score. Image-wise treats every image as an independent instance. Best performances are \textbf{bold}, with the runner-up \underline{underlined}. We execute our method for $n=5$ runs and report mean $\pm$ standard deviation.}
\label{tab:main_detection_results}
\end{table*}

We evaluate our approach both quantitatively and qualitatively and compare it to several state-of-the-art methods.

\subsection{Evaluation Setting}
All models are trained in a semi-supervised fashion, i.e. all training examples are known to be defect-free.
Furthermore, the model knows the camera-to-object grouping, enabling it to reason about an object across views.
Validation is done on the standard test set, containing both normal and large amounts of anomalous samples.
We train one model per object class, and compare to a wide range of state-of-the-art algorithms, similarly trained in a semi-supervised and one-model-per-class fashion.
Comparisons are done to baseline models PaDiM~\cite{padim}, Cflow~\cite{cflow}, SoftPatch~\cite{softpatch}, DeSTSeg~\cite{destseg}, Reverse-Destillation~\cite{rd}, UniAD~\cite{uniad}, PatchCore~\cite{patchcore} and SimpleNet~\cite{simplenet}, whose numbers are taken from the original Real-IAD paper~\cite{realiad}.
These provide a good range across both older and recent state-of-the-art methods as well as a mix of different paradigms, such as density estimation or student-teacher networks.

\subsection{Dataset and Metrics} \emph{Real-IAD}~\cite{realiad} is a large anomaly detection data set, compromised of 30 different object classes, with roughly 3000 defect-free and 1700 anomalous images per class. This is magnitudes larger than previous data sets~\cite{dataset:mvtec, dataset:visa, dataset:btad}. Each object instance contains views from the object sides (i.e. 4 side-views) and one top-view from the bird perspective. Anomalies are grouped into one of eight categories, such as deformations, scratches, missing parts, or foreign objects.

We follow the standard procedures, as proposed by the respective data sets, for evaluation~\cite{dataset:mvtec, realiad}.
The detection may either be measured \emph{image-wise} (i.e.\ each image is treated as an independent instance), or \emph{sample-wise} (multi-view images of the same object instance are aggregated into one sample).
Both are measured using the Area Under the Receiver Operating Curve (AUROC), which measures detection performance independent of any thresholds.
Segmenting the anomalous regions will be evaluated by calculating the AUROC for each pixel, and the AUPRO~\cite{dataset:mvtec}, which is more sophisticated at measuring performance irrespective of the size of anomaly.

\subsection{Implementation Details}
We resize all images to a size of $768\times768$ pixels, and pass them through the extractor $f_F$, which is implemented by an EfficientNet-b5~\cite{Efficientnet} pre-trained on the general purpose ImageNet data set~\cite{imagenet}.
Similar to other methods~\cite{csflow, AST_rudolph} we extract the intermediate features after the $36$-th layer, resulting in $304$ feature maps of size $24\times24$ pixels, which suffices for a good anomaly segmentation performance.
We chain together six coupling blocks for our architecture, with the scale factors clamped using $\alpha = 1.9$ in~\cref{eq:ardizzone}.

Training and optimization is done for $112$ epochs using the AdamW optimizer~\cite{adamw} with a learning rate of $2 \cdot 10^{-4}$, a weight decay of $10^{-5},$ and momentum parameters $\beta_1 = 0.9$ and $\beta_2 = 0.95$.
The ConvBlocks in the $st$-networks work with a hidden feature dimension of $64$, which is brought back up by the later convolutional connections.
The noise $\epsilon$ is sampled with a fixed $c^2 = 0.15$.
We follow the data loading implementation from Real-IAD~\cite{realiad}, leaving out any augmentations that may alter the contents of the images.
Evaluation is done with the efficient and GPU-accelerated computation methods provided by ADEval~\cite{ader}.

\subsection{Anomaly Detection Results}

The main results for the image and sample-wise anomaly detection, averaged across all classes of Real-IAD, can be found in~\cref{tab:main_detection_results}.
More comprehensive results for each of the single classes are in the appendix in~\cref{tab:big_table_samplewise} and~\cref{tab:big_table_imagewise}.

As visible in~\cref{tab:main_detection_results}, we are able to outperform all baselines in the detection tasks.
In the sample-wise prediction task, where we aggregate the anomaly scores across views, we reach a new state-of-the-art with an AUROC of $95.85$.
This is an improvement over the current best-performing model SimpleNet, which sits at a sample-wise AUROC of $94.9$. 
As for the image-wise detection, where each image is treated as an i.i.d. sample, we also slightly outperform all baselines with an AUROC of $90.27$, while the next best competitors SoftPatch and PatchCore both reach $89.4$.

Since pixel-precise and accurate segmentation is not the focus of our method, we slightly lack behind in the segmentation metrics, as we achieve a pixel-wise AUROC of $96.47$ and an AUPRO of $87.91$, while the best competitor RD achieves and AUPRO of $93.8$.
The full results of the segmentation task can be found in the appendix in~\cref{tab:appendix_segmentation}.
Future work could deal with improving the segmentation capabilities of normalizing flow-based anomaly detection methods, which have been known to be improvable, despite their very strong detection performance~\cite{differnet, csflow}.

\begin{figure*}[ht]
    \centering
    \includegraphics[width=0.9\linewidth]{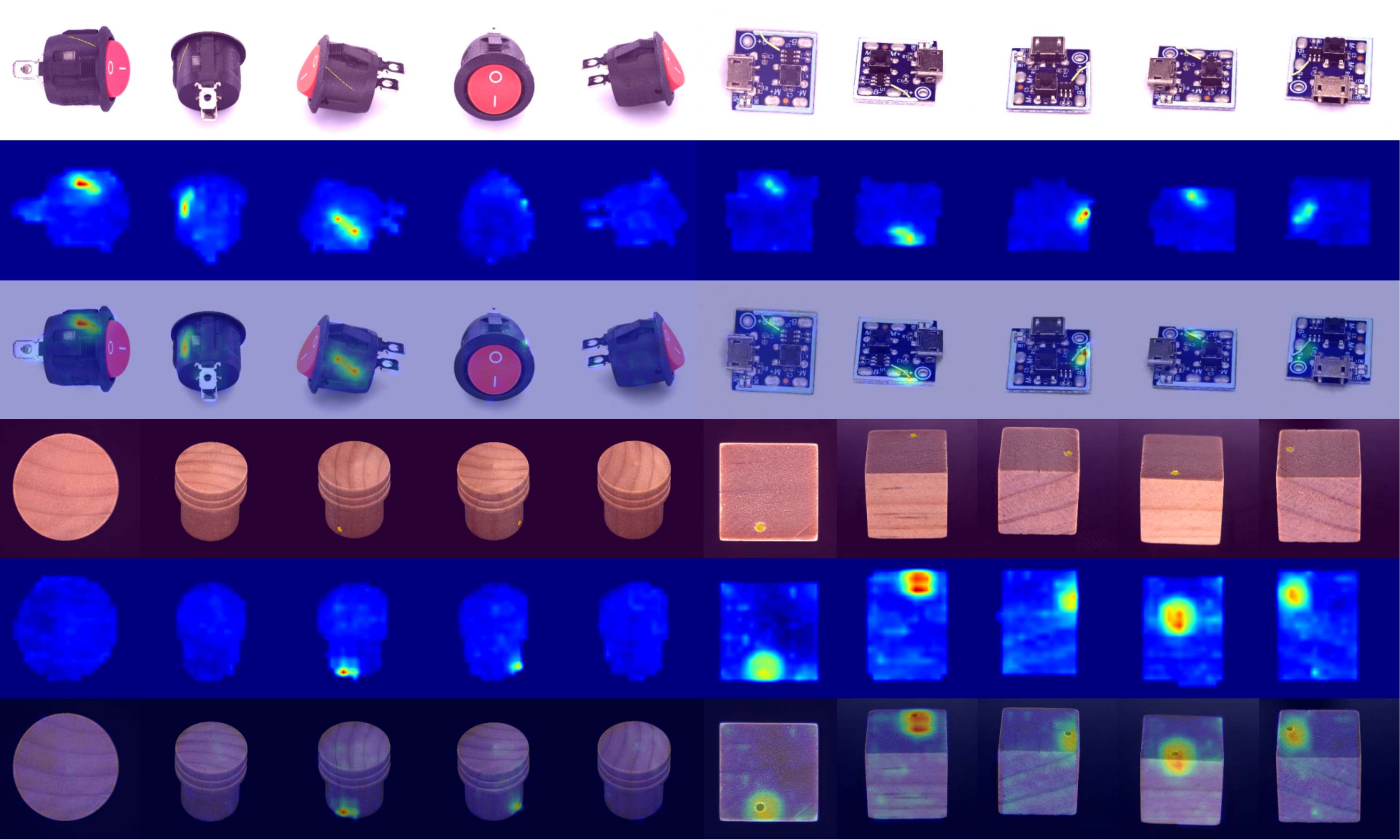}
    \caption{Qualitative results of detecting anomalies in various multi-view poses. For each object, the top row contains the image with anomalies marked. The middle row is the raw anomaly map output by the model. The lower row contains a superimposition of both input image and anomaly map values. The depicted classes are "Switch", "PCB", "Fire Hood", and "Toy Brick".}
    \label{fig:qualtitative_maps}
\end{figure*}

We also provide qualitative results for four different objects, seen from all five viewpoints, in~\cref{fig:qualtitative_maps}.
Here, we can observe that anomalies are being detected in all of the different views of the camera setup, irrespective of their location.
Furthermore, when anomalies are only present in some of the views, their errors do not wrongly carry over to views without anomalies. 
This enables our approach to also perform well in the image-wise setting, where each view is treated independently.

\subsection{Ablation Studies}

We conduct ablation studies on our design choices, including cross-view connections, the background removal, the fusion functions $f_\text{cross}$, and the noise conditioning.

\subsubsection{Impact of Cross-View Connections}

We conduct an ablation study on the effectiveness of the cross-view connections introduced in~\cref{sec:st_network} and visualized in~\cref{fig:crossview_architecture}.
We remove the connections to neighboring and/or to the top-view and re-run our experiments.
The result of this ablation can be found in~\cref{tab:ablation_connections}.
Here, it becomes apparent that removing all cross-view connections leads to a worse sample-wise detection performance.
Further, including any of our proposed cross-view connection blocks improves the sample-wise detection substantially, while slightly deteriorating the image-wise score.
Lastly, the combination of both types of connections, top-view and neighboring-view, leads to the best performance in both metrics, which further motivates our choice of cross-view connections and architecture.

\begin{table}[ht]
    \centering
    \resizebox{\linewidth}{!}{
    \begin{tabular}{c|c||c|c}
        \multicolumn{2}{c||}{Connection Type} & \multicolumn{2}{c}{AUROC ($\uparrow$)}\\
        Top-View & Neighboring & Sample-wise & Image-wise \\\hline\hline
        \texttimes & \texttimes & 94.62 & \underline{90.25} \\
         \checkmark & \texttimes & 95.50 & 90.03 \\
          \texttimes & \checkmark & \underline{95.61} & 90.19 \\
           \checkmark & \checkmark & \textbf{95.85} & \textbf{90.27} \\
    \end{tabular}
    }
    \caption{Ablation on different cross-view neighborhood choices. Best performances are \textbf{bold}, with the runner-up \underline{underlined}.}
    \label{tab:ablation_connections}
\end{table}

\subsubsection{Impact of Background Removal}

Removing the background and calculating the loss only on the foreground object greatly boosts the detection performance.
The image-wise detection is boosted from an AUROC of $85.00$ across all classes in Real-IAD to an AUROC of $90.27$, which is a $6.2\%$ boost in performance.
This may be attributed to the model focusing more clearly on the object. 
Further, it is no longer confused by dust particles or other irritating background factors.
This makes background removal a crucial feature for precise detection performance.

However, it does come at a cost, as the segmentation performance, measured by the Pixel-AUROC metric, drops from $97.29$ to $96.47$ when removing the background information, while AUPRO stays roughly the same.
We argue, that while the masks provided by the MVANet are very precise, they do end up pruning some of the relevant information.
The pixels in the extracted feature space all contain information tied to a larger receptive field.
This may in turn result in removed feature pixels to still contain some valuable information about the object, to which the model does not have access anymore.
Furthermore, masking of anomalous regions, such as missing parts, is introduced as a new, albeit rare, source of error.
Nevertheless, we still suggest to use the background removal, as it guides the model to focus on the important parts of the feature distribution.
Any successes in detection or segmentation that may be attributed to uncontrollable background factors are eliminated after all.

\subsection{Impact of Feature-Fusion Architecture}

We further test the capabilities of other information fusion techniques for our cross-view connections.

\begin{table}[h]
    \centering
    \begin{tabular}{l||c|c|c|c}
        \multicolumn{2}{l}{} & \multicolumn{3}{c}{\leftarrowfill AUROC ($\uparrow$)\rightarrowfill} \\
        \multicolumn{1}{c||}{$f_{\text{cross}}$} & AUPRO & Pixel & Sample & Image  \\\hline\hline
        Self-Attention  & \uline{87.94}          & \uline{96.48}  & 95.19          & 89.63           \\
        Cross-View Att. & \textbf{88.17} & \textbf{96.54} & \uline{95.27}  & \uline{89.95}   \\
        2D Convolution  & 87.91  & 96.47  & \textbf{95.85} & \textbf{90.27} 
    \end{tabular}
    \caption{Comparison of different multi-view feature fusion methods for the cross-view connections $f_\text{cross}$. Best performances are \textbf{bold}, with the runner-up \underline{underlined}.}
    \label{tab:ablation_attention}
\end{table}

All the cross-view connections $f_\text{cross}$ discussed in~\cref{sec:st_network} and~\cref{fig:crossview_architecture} consist of a 2D convolutional layer to finetune the neighboring features.
Here, the convolution is applied to the neighboring view, and summed up together with the original source view, as stated in~\cref{eq:neighbor_sums}.
This mechanism is most easily replaced by an attention mechanism applied to the neighboring view, which is the main building block of the popular vision transformers~\cite{vit}.
For self-attention, instead of a single 2D convolution, we apply a multi-head self-attention layer, layer normalization, and ReLU activation to the neighboring view before summation.
Adding to that, for cross-view attention, we use the source view $\hat{y}^i$ as query and key to the attention mechanism, and the neighboring views $\hat{y}^j \in N\left(\hat{y}^i\right)$ as the value, while keeping the rest of the block the same.

As visible in~\cref{tab:ablation_attention}, cross-view attention beats out self-attention in all metrics. 
However, the simplicity and fast convergence of the 2D convolutions out-performs both attention mechanisms in the detection tasks, while being slightly inferior in the segmentation metrics.
Since attention layers greatly increase both memory requirements and inference speed, we prefer to use the convolutional blocks for deployment in real-life scenarios.
However, for larger data sets, switching to attention may yield more benefit and could be part of future research endeavors~\cite{smalldata_vit}.

\subsection{Impact of Augmentation \& Conditioning}

We study the impact of using the different noise schemes for augmenting the data and conditioning our normalizing flow.
The results for anomaly detection can be found in~\cref{tab:ablation_noise}.
\method~ samples noise similar to SimpleNet~\cite{simplenet}, with vectors $\epsilon \sim \mathcal{N}\left(0,0.15\cdot I\right)$ which are added to the input features, to regularize training.
We also test sampling noise uniformly as $\epsilon \sim U\left(0,1\right)$.
Similar to SoftFlow~\cite{softflow}, which originally proposed conditioning normalizing flows with noise, we also test their sampling strategy. It involves drawing a vector $c \sim U\left(0,1\right)$ and then sampling noise as $\epsilon \sim \mathcal{N}\left(0, c^2\cdot I\right)$.
Compared to SimpleNet, this results in more variance in the noise distribution.

As visible in~\cref{tab:ablation_noise}, the different noise conditioning schemes only slightly influence the segmentation metrics AUPRO and Pixel-AUROC.
As for the detection tasks, which is our focus, there are slight advantages for both the SoftFlow and SimpleNet schemes for the image and sample-wise AUROC respectively. 
In the end, we propose to use the conditioning scheme proposed by SimpleNet, resulting in a slightly boosted sample-wise performance, as well as a relatively strong performance in all other metrics.

\begin{table}[ht]
    \centering
    \begin{tabular}{l||c|c|c|c}
        \multicolumn{2}{l}{} & \multicolumn{3}{c}{\leftarrowfill AUROC ($\uparrow$)\rightarrowfill} \\
        Noise-type & AUPRO & Pixel & Sample & Image  \\\hline\hline
        No Noise  & \textbf{87.92} & \textbf{\uline{96.47}} & \uline{95.84}  & 90.25          \\
Uniform   & 87.67          & 96.38                  & 95.54          & 90.23          \\
SoftFlow~\cite{softflow}  & 87.89          & \textbf{\uline{96.47}} & 95.77          & \textbf{90.41} \\
SimpleNet~\cite{simplenet} & \uline{87.91}  & \textbf{\uline{96.47}} & \textbf{95.85} & \uline{90.27}   
    \end{tabular}
    \caption{Comparison on the noise conditioning, with \method~ using the SimpleNet-esque method. Best performances are \textbf{bold}, with the runner-up \underline{underlined}.}
    \label{tab:ablation_noise}
\end{table}

\section{Conclusion}\label{sec:conclusion}

In this paper, we proposed a novel normalizing flow-based architecture to perform anomaly detection in multi-view image data.
We have designed a new architecture and training scheme tailored to multi-view setups.
This allows the flow to share its information across views, boosting the performance in both image and sample-wise anomaly detection.

Future work could look towards extending to data sets with less rigid camera setups, where dozens of views of an object may be available~\cite{splatpose, dataset:pad}.
Encoding similar architectural priors into other tasks, such as logical anomaly detection~\cite{dataset:mvtecloco}, may be another interesting future avenue.

\small{\paragraph{Acknowledgements.} This work was supported by the Federal Ministry of Education and Research (BMBF), Germany, under the AI service center KISSKI (grant no. 01IS22093C), the MWK of Lower Sachsony within Hybrint (VWZN4219), the Deutsche Forschungsgemeinschaft (DFG) under Germany’s Excellence Strategy within the Cluster of Excellence PhoenixD (EXC2122), the European Union  under grant agreement no. 101136006 – XTREME.}

\clearpage

{
    \small
    \bibliographystyle{ieeenat_fullname}
    \bibliography{main}

\begin{thebibliography}{71}
\providecommand{\natexlab}[1]{#1}
\providecommand{\url}[1]{\texttt{#1}}
\expandafter\ifx\csname urlstyle\endcsname\relax
  \providecommand{\doi}[1]{doi: #1}\else
  \providecommand{\doi}{doi: \begingroup \urlstyle{rm}\Url}\fi

\bibitem[Akcay et~al.(2019)Akcay, Atapour-Abarghouei, and Breckon]{ganomaly}
Samet Akcay, Amir Atapour-Abarghouei, and Toby~P Breckon.
\newblock Ganomaly: Semi-supervised anomaly detection via adversarial training.
\newblock In \emph{Computer Vision--ACCV 2018: 14th Asian Conference on Computer Vision, Perth, Australia, December 2--6, 2018, Revised Selected Papers, Part III 14}, pages 622--637. Springer, 2019.

\bibitem[Ardizzone et~al.(2019)Ardizzone, L{\"{u}}th, Kruse, Rother, and K{\"{o}}the]{ardizzone_nf}
Lynton Ardizzone, Carsten~T. L{\"{u}}th, Jakob Kruse, Carsten Rother, and Ullrich K{\"{o}}the.
\newblock Guided image generation with conditional invertible neural networks.
\newblock \emph{CoRR}, abs/1907.02392, 2019.

\bibitem[Bae et~al.(2023)Bae, Lee, and Kim]{pni}
Jaehyeok Bae, Jae{-}Han Lee, and Seyun Kim.
\newblock {PNI:} industrial anomaly detection using position and neighborhood information.
\newblock In \emph{{IEEE/CVF} International Conference on Computer Vision, {ICCV} 2023, Paris, France, October 1-6, 2023}, pages 6350--6360. {IEEE}, 2023.

\bibitem[Batzner et~al.(2024)Batzner, Heckler, and K\"onig]{efficientAD}
Kilian Batzner, Lars Heckler, and Rebecca K\"onig.
\newblock Efficientad: Accurate visual anomaly detection at millisecond-level latencies.
\newblock In \emph{Proceedings of the IEEE/CVF Winter Conference on Applications of Computer Vision (WACV)}, pages 128--138, 2024.

\bibitem[Bergmann et~al.(2019{\natexlab{a}})Bergmann, Fauser, Sattlegger, and Steger]{dataset:mvtec}
Paul Bergmann, Michael Fauser, David Sattlegger, and Carsten Steger.
\newblock Mvtec ad--a comprehensive real-world dataset for unsupervised anomaly detection.
\newblock In \emph{Proceedings of the IEEE/CVF conference on computer vision and pattern recognition}, pages 9592--9600, 2019{\natexlab{a}}.

\bibitem[Bergmann et~al.(2019{\natexlab{b}})Bergmann, L{\"{o}}we, Fauser, Sattlegger, and Steger]{ae_ad}
Paul Bergmann, Sindy L{\"{o}}we, Michael Fauser, David Sattlegger, and Carsten Steger.
\newblock Improving unsupervised defect segmentation by applying structural similarity to autoencoders.
\newblock In \emph{Proceedings of the 14th International Joint Conference on Computer Vision, Imaging and Computer Graphics Theory and Applications, {VISIGRAPP} 2019, Volume 5: VISAPP, Prague, Czech Republic, February 25-27, 2019}, pages 372--380, 2019{\natexlab{b}}.

\bibitem[Bergmann et~al.(2022{\natexlab{a}})Bergmann, Batzner, Fauser, Sattlegger, and Steger]{dataset:mvtecloco}
Paul Bergmann, Kilian Batzner, Michael Fauser, David Sattlegger, and Carsten Steger.
\newblock Beyond dents and scratches: Logical constraints in unsupervised anomaly detection and localization.
\newblock \emph{International Journal of Computer Vision}, 130\penalty0 (4):\penalty0 947--969, 2022{\natexlab{a}}.

\bibitem[Bergmann et~al.(2022{\natexlab{b}})Bergmann, Jin, Sattlegger, and Steger]{dataset:mvtec3d}
Paul Bergmann, Xin Jin, David Sattlegger, and Carsten Steger.
\newblock The mvtec 3d-ad dataset for unsupervised 3d anomaly detection and localization.
\newblock In \emph{Proceedings of the 17th International Joint Conference on Computer Vision, Imaging and Computer Graphics Theory and Applications, {VISIGRAPP} 2022, Volume 5: VISAPP, Online Streaming, February 6-8, 2022}, pages 202--213. {SCITEPRESS}, 2022{\natexlab{b}}.

\bibitem[Bonfiglioli et~al.(2022)Bonfiglioli, Toschi, Silvestri, Fioraio, and De~Gregorio]{dataset:eycandies}
Luca Bonfiglioli, Marco Toschi, Davide Silvestri, Nicola Fioraio, and Daniele De~Gregorio.
\newblock The eyecandies dataset for unsupervised multimodal anomaly detection and localization.
\newblock In \emph{Proceedings of the Asian Conference on Computer Vision}, pages 3586--3602, 2022.

\bibitem[Brockmann et~al.(2024)Brockmann, Rudolph, Rosenhahn, and Wandt]{vorausAD}
Jan~Thie{\ss} Brockmann, Marco Rudolph, Bodo Rosenhahn, and Bastian Wandt.
\newblock The voraus-ad dataset for anomaly detection in robot applications.
\newblock \emph{{IEEE} Trans. Robotics}, 40:\penalty0 438--451, 2024.

\bibitem[Cohen and Hoshen(2020)]{SPADE}
Niv Cohen and Yedid Hoshen.
\newblock Sub-image anomaly detection with deep pyramid correspondences.
\newblock \emph{arXiv preprint arXiv:2005.02357}, 2020.

\bibitem[Defard et~al.(2021)Defard, Setkov, Loesch, and Audigier]{padim}
Thomas Defard, Aleksandr Setkov, Angelique Loesch, and Romaric Audigier.
\newblock Padim: a patch distribution modeling framework for anomaly detection and localization.
\newblock In \emph{International Conference on Pattern Recognition}, pages 475--489. Springer, 2021.

\bibitem[Deng and Li(2022)]{rd}
Hanqiu Deng and Xingyu Li.
\newblock Anomaly detection via reverse distillation from one-class embedding.
\newblock In \emph{Proceedings of the IEEE/CVF Conference on Computer Vision and Pattern Recognition}, pages 9737--9746, 2022.

\bibitem[Deng et~al.(2009)Deng, Dong, Socher, Li, Li, and Fei-Fei]{imagenet}
Jia Deng, Wei Dong, Richard Socher, Li-Jia Li, Kai Li, and Li Fei-Fei.
\newblock Imagenet: A large-scale hierarchical image database.
\newblock In \emph{2009 IEEE conference on computer vision and pattern recognition}, pages 248--255. Ieee, 2009.

\bibitem[Dinh et~al.(2015)Dinh, Krueger, and Bengio]{NF_NICE}
Laurent Dinh, David Krueger, and Yoshua Bengio.
\newblock {NICE:} non-linear independent components estimation.
\newblock In \emph{3rd International Conference on Learning Representations, {ICLR} 2015, San Diego, CA, USA, May 7-9, 2015, Workshop Track Proceedings}, 2015.

\bibitem[Dinh et~al.(2017)Dinh, Sohl{-}Dickstein, and Bengio]{RealNVP}
Laurent Dinh, Jascha Sohl{-}Dickstein, and Samy Bengio.
\newblock Density estimation using real {NVP}.
\newblock In \emph{5th International Conference on Learning Representations, {ICLR} 2017, Toulon, France, April 24-26, 2017, Conference Track Proceedings}. OpenReview.net, 2017.

\bibitem[Dosovitskiy et~al.(2021)Dosovitskiy, Beyer, Kolesnikov, Weissenborn, Zhai, Unterthiner, Dehghani, Minderer, Heigold, Gelly, Uszkoreit, and Houlsby]{vit}
Alexey Dosovitskiy, Lucas Beyer, Alexander Kolesnikov, Dirk Weissenborn, Xiaohua Zhai, Thomas Unterthiner, Mostafa Dehghani, Matthias Minderer, Georg Heigold, Sylvain Gelly, Jakob Uszkoreit, and Neil Houlsby.
\newblock An image is worth 16x16 words: Transformers for image recognition at scale.
\newblock In \emph{9th International Conference on Learning Representations, {ICLR} 2021, Virtual Event, Austria, May 3-7, 2021}. OpenReview.net, 2021.

\bibitem[Durkan et~al.(2019)Durkan, Bekasov, Murray, and Papamakarios]{neural_spline_flows}
Conor Durkan, Artur Bekasov, Iain Murray, and George Papamakarios.
\newblock Neural spline flows.
\newblock In \emph{Advances in Neural Information Processing Systems 32: Annual Conference on Neural Information Processing Systems 2019, NeurIPS 2019, December 8-14, 2019, Vancouver, BC, Canada}, pages 7509--7520, 2019.

\bibitem[Germain et~al.(2015)Germain, Gregor, Murray, and Larochelle]{NF:MADE}
Mathieu Germain, Karol Gregor, Iain Murray, and Hugo Larochelle.
\newblock {MADE:} masked autoencoder for distribution estimation.
\newblock In \emph{Proceedings of the 32nd International Conference on Machine Learning, {ICML} 2015, Lille, France, 6-11 July 2015}, pages 881--889. JMLR.org, 2015.

\bibitem[Gu et~al.(2024)Gu, Zhu, Zhu, Chen, Tang, and Wang]{AnomalyGpt}
Zhaopeng Gu, Bingke Zhu, Guibo Zhu, Yingying Chen, Ming Tang, and Jinqiao Wang.
\newblock Anomalygpt: Detecting industrial anomalies using large vision-language models.
\newblock In \emph{Thirty-Eighth {AAAI} Conference on Artificial Intelligence, {AAAI} 2024}, pages 1932--1940. {AAAI} Press, 2024.

\bibitem[Gudovskiy et~al.(2022)Gudovskiy, Ishizaka, and Kozuka]{cflow}
Denis Gudovskiy, Shun Ishizaka, and Kazuki Kozuka.
\newblock Cflow-ad: Real-time unsupervised anomaly detection with localization via conditional normalizing flows.
\newblock In \emph{Proceedings of the IEEE/CVF Winter Conference on Applications of Computer Vision}, pages 98--107, 2022.

\bibitem[Heckler et~al.(2023)Heckler, K{\"{o}}nig, and Bergmann]{heckler_features}
Lars Heckler, Rebecca K{\"{o}}nig, and Paul Bergmann.
\newblock Exploring the importance of pretrained feature extractors for unsupervised anomaly detection and localization.
\newblock In \emph{{IEEE/CVF} Conference on Computer Vision and Pattern Recognition, {CVPR} 2023 - Workshops, Vancouver, BC, Canada, June 17-24, 2023}, pages 2917--2926. {IEEE}, 2023.

\bibitem[Ho et~al.(2019)Ho, Chen, Srinivas, Duan, and Abbeel]{flow_plusplus}
Jonathan Ho, Xi Chen, Aravind Srinivas, Yan Duan, and Pieter Abbeel.
\newblock Flow++: Improving flow-based generative models with variational dequantization and architecture design.
\newblock In \emph{Proceedings of the 36th International Conference on Machine Learning, {ICML} 2019, 9-15 June 2019, Long Beach, California, {USA}}, pages 2722--2730. {PMLR}, 2019.

\bibitem[Ho et~al.(2020)Ho, Jain, and Abbeel]{ddpm_probabilistic}
Jonathan Ho, Ajay Jain, and Pieter Abbeel.
\newblock Denoising diffusion probabilistic models.
\newblock In \emph{Advances in Neural Information Processing Systems 33: Annual Conference on Neural Information Processing Systems 2020, NeurIPS 2020, December 6-12, 2020, virtual}, 2020.

\bibitem[Jeong et~al.(2023)Jeong, Zou, Kim, Zhang, Ravichandran, and Dabeer]{WINClip}
Jongheon Jeong, Yang Zou, Taewan Kim, Dongqing Zhang, Avinash Ravichandran, and Onkar Dabeer.
\newblock Winclip: Zero-/few-shot anomaly classification and segmentation.
\newblock In \emph{{IEEE/CVF} Conference on Computer Vision and Pattern Recognition, {CVPR} 2023, Vancouver, BC, Canada, June 17-24, 2023}, pages 19606--19616. {IEEE}, 2023.

\bibitem[Jiang et~al.(2022)Jiang, Liu, Wang, Nie, Wu, Liu, Wang, and Zheng]{softpatch}
Xi Jiang, Jianlin Liu, Jinbao Wang, Qiang Nie, Kai Wu, Yong Liu, Chengjie Wang, and Feng Zheng.
\newblock Softpatch: Unsupervised anomaly detection with noisy data.
\newblock \emph{Advances in Neural Information Processing Systems}, 35:\penalty0 15433--15445, 2022.

\bibitem[Kim et~al.(2020)Kim, Lee, Kang, Lee, and Kim]{softflow}
Hyeongju Kim, Hyeonseung Lee, Woo~Hyun Kang, Joun~Yeop Lee, and Nam~Soo Kim.
\newblock Softflow: Probabilistic framework for normalizing flow on manifolds.
\newblock In \emph{Advances in Neural Information Processing Systems 33: Annual Conference on Neural Information Processing Systems 2020, NeurIPS 2020, December 6-12, 2020, virtual}, 2020.

\bibitem[Kingma and Dhariwal(2018)]{glow_openai}
Diederik~P. Kingma and Prafulla Dhariwal.
\newblock Glow: Generative flow with invertible 1x1 convolutions.
\newblock In \emph{Advances in Neural Information Processing Systems 31: Annual Conference on Neural Information Processing Systems 2018, NeurIPS 2018, December 3-8, 2018, Montr{\'{e}}al, Canada}, pages 10236--10245, 2018.

\bibitem[Kingma et~al.(2016)Kingma, Salimans, J{\'{o}}zefowicz, Chen, Sutskever, and Welling]{nf:IAF}
Diederik~P. Kingma, Tim Salimans, Rafal J{\'{o}}zefowicz, Xi Chen, Ilya Sutskever, and Max Welling.
\newblock Improving variational autoencoders with inverse autoregressive flow.
\newblock In \emph{Advances in Neural Information Processing Systems 29: Annual Conference on Neural Information Processing Systems 2016, December 5-10, 2016, Barcelona, Spain}, pages 4736--4744, 2016.

\bibitem[Kirichenko et~al.(2020)Kirichenko, Izmailov, and Wilson]{nf_kirichenko_ood}
Polina Kirichenko, Pavel Izmailov, and Andrew~Gordon Wilson.
\newblock Why normalizing flows fail to detect out-of-distribution data.
\newblock In \emph{Advances in Neural Information Processing Systems 33: Annual Conference on Neural Information Processing Systems 2020, NeurIPS 2020, December 6-12, 2020, virtual}, 2020.

\bibitem[Kruse et~al.(2024)Kruse, Rudolph, Woiwode, and Rosenhahn]{splatpose}
Mathis Kruse, Marco Rudolph, Dominik Woiwode, and Bodo Rosenhahn.
\newblock Splatpose {\&} detect: Pose-agnostic 3d anomaly detection.
\newblock In \emph{{IEEE/CVF} Conference on Computer Vision and Pattern Recognition, {CVPR} 2024 - Workshops, Seattle, WA, USA, June 17-18, 2024}, pages 3950--3960. {IEEE}, 2024.

\bibitem[Lee et~al.(2022)Lee, Lee, and Song]{cfa}
Sungwook Lee, Seunghyun Lee, and Byung~Cheol Song.
\newblock {CFA:} coupled-hypersphere-based feature adaptation for target-oriented anomaly localization.
\newblock \emph{{IEEE} Access}, 10:\penalty0 78446--78454, 2022.

\bibitem[Lei et~al.(2023)Lei, Hu, Wang, and Liu]{pyramidflow}
Jiarui Lei, Xiaobo Hu, Yue Wang, and Dong Liu.
\newblock Pyramidflow: High-resolution defect contrastive localization using pyramid normalizing flow.
\newblock In \emph{Proceedings of the IEEE/CVF Conference on Computer Vision and Pattern Recognition}, pages 14143--14152, 2023.

\bibitem[Li et~al.(2021)Li, Sohn, Yoon, and Pfister]{cutpaste}
Chun-Liang Li, Kihyuk Sohn, Jinsung Yoon, and Tomas Pfister.
\newblock Cutpaste: Self-supervised learning for anomaly detection and localization.
\newblock In \emph{Proceedings of the IEEE/CVF conference on computer vision and pattern recognition}, pages 9664--9674, 2021.

\bibitem[Li et~al.(2024)Li, Xu, Gu, Zheng, Gao, and Wu]{dataset:anomaly_shapenet}
Wenqiao Li, Xiaohao Xu, Yao Gu, Bozhong Zheng, Shenghua Gao, and Yingna Wu.
\newblock Towards scalable 3d anomaly detection and localization: {A} benchmark via 3d anomaly synthesis and {A} self-supervised learning network.
\newblock In \emph{{IEEE/CVF} Conference on Computer Vision and Pattern Recognition, {CVPR} 2024, Seattle, WA, USA, June 16-22, 2024}, pages 22207--22216. {IEEE}, 2024.

\bibitem[Liang et~al.(2023)Liang, Zhang, Zhao, Wu, Liu, and Pan]{ocrgan}
Yufei Liang, Jiangning Zhang, Shiwei Zhao, Runze Wu, Yong Liu, and Shuwen Pan.
\newblock Omni-frequency channel-selection representations for unsupervised anomaly detection.
\newblock \emph{IEEE Transactions on Image Processing}, 2023.

\bibitem[Liu et~al.(2023{\natexlab{a}})Liu, Xie, Chen, Li, Wang, Liu, Wang, and Zheng]{dataset:real3d}
Jiaqi Liu, Guoyang Xie, Ruitao Chen, Xinpeng Li, Jinbao Wang, Yong Liu, Chengjie Wang, and Feng Zheng.
\newblock Real3d-ad: {A} dataset of point cloud anomaly detection.
\newblock In \emph{Advances in Neural Information Processing Systems 36: Annual Conference on Neural Information Processing Systems 2023, NeurIPS 2023, New Orleans, LA, USA, December 10 - 16, 2023}, 2023{\natexlab{a}}.

\bibitem[Liu et~al.(2023{\natexlab{b}})Liu, Zhou, Xu, and Wang]{simplenet}
Zhikang Liu, Yiming Zhou, Yuansheng Xu, and Zilei Wang.
\newblock Simplenet: A simple network for image anomaly detection and localization.
\newblock In \emph{Proceedings of the IEEE/CVF Conference on Computer Vision and Pattern Recognition}, pages 20402--20411, 2023{\natexlab{b}}.

\bibitem[Loshchilov and Hutter(2019)]{adamw}
Ilya Loshchilov and Frank Hutter.
\newblock Decoupled weight decay regularization.
\newblock In \emph{7th International Conference on Learning Representations, {ICLR} 2019, New Orleans, LA, USA, May 6-9, 2019}. OpenReview.net, 2019.

\bibitem[Lu et~al.(2023)Lu, Wu, Tian, Wang, Chen, Liu, and Hu]{multiclass_vq}
Ruiying Lu, Yujie Wu, Long Tian, Dongsheng Wang, Bo Chen, Xiyang Liu, and Ruimin Hu.
\newblock Hierarchical vector quantized transformer for multi-class unsupervised anomaly detection.
\newblock In \emph{Advances in Neural Information Processing Systems 36: Annual Conference on Neural Information Processing Systems 2023, NeurIPS 2023, New Orleans, LA, USA, December 10 - 16, 2023}, 2023.

\bibitem[Lu et~al.(2022)Lu, Xie, Liu, and Zhang]{smalldata_vit}
Zhiying Lu, Hongtao Xie, Chuanbin Liu, and Yongdong Zhang.
\newblock Bridging the gap between vision transformers and convolutional neural networks on small datasets.
\newblock In \emph{Advances in Neural Information Processing Systems 35: Annual Conference on Neural Information Processing Systems 2022, NeurIPS 2022, New Orleans, LA, USA, November 28 - December 9, 2022}, 2022.

\bibitem[Mishra et~al.(2021)Mishra, Verk, Fornasier, Piciarelli, and Foresti]{dataset:btad}
Pankaj Mishra, Riccardo Verk, Daniele Fornasier, Claudio Piciarelli, and Gian~Luca Foresti.
\newblock Vt-adl: A vision transformer network for image anomaly detection and localization.
\newblock In \emph{2021 IEEE 30th International Symposium on Industrial Electronics (ISIE)}, pages 01--06. IEEE, 2021.

\bibitem[Pang et~al.(2019)Pang, Shen, and van~den Hengel]{devnet}
Guansong Pang, Chunhua Shen, and Anton van~den Hengel.
\newblock Deep anomaly detection with deviation networks.
\newblock In \emph{Proceedings of the 25th {ACM} {SIGKDD} International Conference on Knowledge Discovery {\&} Data Mining, {KDD} 2019, Anchorage, AK, USA, August 4-8, 2019}, pages 353--362. {ACM}, 2019.

\bibitem[Papamakarios et~al.(2017)Papamakarios, Murray, and Pavlakou]{NF:MAF}
George Papamakarios, Iain Murray, and Theo Pavlakou.
\newblock Masked autoregressive flow for density estimation.
\newblock In \emph{Advances in Neural Information Processing Systems 30: Annual Conference on Neural Information Processing Systems 2017, December 4-9, 2017, Long Beach, CA, {USA}}, pages 2338--2347, 2017.

\bibitem[Radford et~al.(2021)Radford, Kim, Hallacy, Ramesh, Goh, Agarwal, Sastry, Askell, Mishkin, Clark, Krueger, and Sutskever]{CLIP}
Alec Radford, Jong~Wook Kim, Chris Hallacy, Aditya Ramesh, Gabriel Goh, Sandhini Agarwal, Girish Sastry, Amanda Askell, Pamela Mishkin, Jack Clark, Gretchen Krueger, and Ilya Sutskever.
\newblock Learning transferable visual models from natural language supervision.
\newblock In \emph{Proceedings of the 38th International Conference on Machine Learning, {ICML} 2021, 18-24 July 2021, Virtual Event}, pages 8748--8763. {PMLR}, 2021.

\bibitem[Rezende and Mohamed(2015)]{rezende_variational_nf}
Danilo~Jimenez Rezende and Shakir Mohamed.
\newblock Variational inference with normalizing flows.
\newblock In \emph{Proceedings of the 32nd International Conference on Machine Learning, {ICML} 2015, Lille, France, 6-11 July 2015}, pages 1530--1538. JMLR.org, 2015.

\bibitem[Rosenhahn and Hirche(2024)]{nf_quantum}
Bodo Rosenhahn and Christoph Hirche.
\newblock Quantum normalizing flows for anomaly detection.
\newblock \emph{Physical Review A}, 110:\penalty0 022443, 2024.

\bibitem[Roth et~al.(2022)Roth, Pemula, Zepeda, Sch{\"o}lkopf, Brox, and Gehler]{patchcore}
Karsten Roth, Latha Pemula, Joaquin Zepeda, Bernhard Sch{\"o}lkopf, Thomas Brox, and Peter Gehler.
\newblock Towards total recall in industrial anomaly detection.
\newblock In \emph{Proceedings of the IEEE/CVF Conference on Computer Vision and Pattern Recognition}, pages 14318--14328, 2022.

\bibitem[Rudolph et~al.(2021)Rudolph, Wandt, and Rosenhahn]{differnet}
Marco Rudolph, Bastian Wandt, and Bodo Rosenhahn.
\newblock Same same but differnet: Semi-supervised defect detection with normalizing flows.
\newblock In \emph{Proceedings of the IEEE/CVF winter conference on applications of computer vision}, pages 1907--1916, 2021.

\bibitem[Rudolph et~al.(2022)Rudolph, Wehrbein, Rosenhahn, and Wandt]{csflow}
Marco Rudolph, Tom Wehrbein, Bodo Rosenhahn, and Bastian Wandt.
\newblock Fully convolutional cross-scale-flows for image-based defect detection.
\newblock In \emph{Proceedings of the IEEE/CVF Winter Conference on Applications of Computer Vision}, pages 1088--1097, 2022.

\bibitem[Rudolph et~al.(2023)Rudolph, Wehrbein, Rosenhahn, and Wandt]{AST_rudolph}
Marco Rudolph, Tom Wehrbein, Bodo Rosenhahn, and Bastian Wandt.
\newblock Asymmetric student-teacher networks for industrial anomaly detection.
\newblock In \emph{WACV}, pages 2591--2601. IEEE, 2023.

\bibitem[Schlegl et~al.(2017)Schlegl, Seeb{\"o}ck, Waldstein, Schmidt-Erfurth, and Langs]{anogan}
Thomas Schlegl, Philipp Seeb{\"o}ck, Sebastian~M Waldstein, Ursula Schmidt-Erfurth, and Georg Langs.
\newblock Unsupervised anomaly detection with generative adversarial networks to guide marker discovery.
\newblock In \emph{International conference on information processing in medical imaging}, pages 146--157. Springer, 2017.

\bibitem[Tan and Le(2019)]{Efficientnet}
Mingxing Tan and Quoc~V. Le.
\newblock Efficientnet: Rethinking model scaling for convolutional neural networks.
\newblock In \emph{Proceedings of the 36th International Conference on Machine Learning, {ICML} 2019, 9-15 June 2019, Long Beach, California, {USA}}, pages 6105--6114. {PMLR}, 2019.

\bibitem[Wang et~al.(2024{\natexlab{a}})Wang, Zhu, Gao, Gan, Zhang, Gu, Qian, Chen, and Ma]{realiad}
Chengjie Wang, Wenbing Zhu, Bin{-}Bin Gao, Zhenye Gan, Jiangning Zhang, Zhihao Gu, Shuguang Qian, Mingang Chen, and Lizhuang Ma.
\newblock Real-iad: {A} real-world multi-view dataset for benchmarking versatile industrial anomaly detection.
\newblock In \emph{{IEEE/CVF} Conference on Computer Vision and Pattern Recognition, {CVPR} 2024, Seattle, WA, USA, June 16-22, 2024}, pages 22883--22892. {IEEE}, 2024{\natexlab{a}}.

\bibitem[Wang et~al.(2024{\natexlab{b}})Wang, Bai, Yu, Zhao, and Xiao]{cvpr24_nf_segmentation}
Xiaoyang Wang, Huihui Bai, Limin Yu, Yao Zhao, and Jimin Xiao.
\newblock Towards the uncharted: Density-descending feature perturbation for semi-supervised semantic segmentation.
\newblock In \emph{{IEEE/CVF} Conference on Computer Vision and Pattern Recognition, {CVPR} 2024, Seattle, WA, USA, June 16-22, 2024}, pages 3303--3312. {IEEE}, 2024{\natexlab{b}}.

\bibitem[Wehrbein et~al.(2021)Wehrbein, Rudolph, Rosenhahn, and Wandt]{pose_nf1}
Tom Wehrbein, Marco Rudolph, Bodo Rosenhahn, and Bastian Wandt.
\newblock Probabilistic monocular 3d human pose estimation with normalizing flows.
\newblock In \emph{2021 {IEEE/CVF} International Conference on Computer Vision, {ICCV} 2021, Montreal, QC, Canada, October 10-17, 2021}, pages 11179--11188. {IEEE}, 2021.

\bibitem[Wehrbein et~al.(2025)Wehrbein, Rudolph, Rosenhahn, and Wandt]{pose_nf2}
Tom Wehrbein, Marco Rudolph, Bodo Rosenhahn, and Bastian Wandt.
\newblock Utilizing uncertainty in 2d pose detectors for probabilistic 3d human mesh recovery.
\newblock In \emph{Proceedings of the IEEE/CVF Winter Conference on Applications of Computer Vision}, 2025.

\bibitem[You et~al.(2022)You, Cui, Shen, Yang, Lu, Zheng, and Le]{uniad}
Zhiyuan You, Lei Cui, Yujun Shen, Kai Yang, Xin Lu, Yu Zheng, and Xinyi Le.
\newblock A unified model for multi-class anomaly detection.
\newblock \emph{Advances in Neural Information Processing Systems}, 35:\penalty0 4571--4584, 2022.

\bibitem[Yu et~al.(2021)Yu, Zheng, Wang, Li, Wu, Zhao, and Wu]{fastflow}
Jiawei Yu, Ye Zheng, Xiang Wang, Wei Li, Yushuang Wu, Rui Zhao, and Liwei Wu.
\newblock Fastflow: Unsupervised anomaly detection and localization via 2d normalizing flows.
\newblock \emph{arXiv preprint arXiv:2111.07677}, 2021.

\bibitem[Yu et~al.(2024)Yu, Zhao, Pang, Zhang, and Lu]{MVANet}
Qian Yu, Xiaoqi Zhao, Youwei Pang, Lihe Zhang, and Huchuan Lu.
\newblock Multi-view aggregation network for dichotomous image segmentation.
\newblock In \emph{Proceedings of the IEEE/CVF Conference on Computer Vision and Pattern Recognition (CVPR)}, pages 3921--3930, 2024.

\bibitem[Zavrtanik et~al.(2021)Zavrtanik, Kristan, and Sko{\v{c}}aj]{draem}
Vitjan Zavrtanik, Matej Kristan, and Danijel Sko{\v{c}}aj.
\newblock Draem-a discriminatively trained reconstruction embedding for surface anomaly detection.
\newblock In \emph{Proceedings of the IEEE/CVF International Conference on Computer Vision}, pages 8330--8339, 2021.

\bibitem[Zhang et~al.(2021)Zhang, Cui, Hung, and Lu]{defectgan}
Gongjie Zhang, Kaiwen Cui, Tzu-Yi Hung, and Shijian Lu.
\newblock Defect-gan: High-fidelity defect synthesis for automated defect inspection.
\newblock In \emph{Proceedings of the IEEE/CVF Winter Conference on Applications of Computer Vision}, pages 2524--2534, 2021.

\bibitem[Zhang et~al.(2024{\natexlab{a}})Zhang, He, Gan, He, Cai, Xue, Wang, Wang, Xie, and Liu]{ader}
Jiangning Zhang, Haoyang He, Zhenye Gan, Qingdong He, Yuxuan Cai, Zhucun Xue, Yabiao Wang, Chengjie Wang, Lei Xie, and Yong Liu.
\newblock Ader: A comprehensive benchmark for multi-class visual anomaly detection.
\newblock \emph{arXiv preprint arXiv:2406.03262}, 2024{\natexlab{a}}.

\bibitem[Zhang et~al.(2023)Zhang, Li, Li, Huang, Shan, and Chen]{destseg}
Xuan Zhang, Shiyu Li, Xi Li, Ping Huang, Jiulong Shan, and Ting Chen.
\newblock Destseg: Segmentation guided denoising student-teacher for anomaly detection.
\newblock In \emph{Proceedings of the IEEE/CVF Conference on Computer Vision and Pattern Recognition}, pages 3914--3923, 2023.

\bibitem[Zhang et~al.(2024{\natexlab{b}})Zhang, Xu, and Zhou]{realnet}
Ximiao Zhang, Min Xu, and Xiuzhuang Zhou.
\newblock Realnet: {A} feature selection network with realistic synthetic anomaly for anomaly detection.
\newblock In \emph{{IEEE/CVF} Conference on Computer Vision and Pattern Recognition, {CVPR} 2024, Seattle, WA, USA, June 16-22, 2024}, pages 16699--16708. {IEEE}, 2024{\natexlab{b}}.

\bibitem[Zhao(2023)]{OmniAL}
Ying Zhao.
\newblock Omnial: {A} unified {CNN} framework for unsupervised anomaly localization.
\newblock In \emph{{IEEE/CVF} Conference on Computer Vision and Pattern Recognition, {CVPR} 2023, Vancouver, BC, Canada, June 17-24, 2023}, pages 3924--3933. {IEEE}, 2023.

\bibitem[Zhou and Paffenroth(2017)]{dae}
Chong Zhou and Randy~C Paffenroth.
\newblock Anomaly detection with robust deep autoencoders.
\newblock In \emph{Proceedings of the 23rd ACM SIGKDD international conference on knowledge discovery and data mining}, pages 665--674, 2017.

\bibitem[Zhou et~al.(2024{\natexlab{a}})Zhou, Cao, Kim, Zhao, Dong, Ting, and Zhu]{dataset:RAD}
Kaichen Zhou, Yang Cao, Taewhan Kim, Hao Zhao, Hao Dong, Kai~Ming Ting, and Ye Zhu.
\newblock {RAD:} {A} dataset and benchmark for real-life anomaly detection with robotic observations.
\newblock \emph{CoRR}, abs/2410.00713, 2024{\natexlab{a}}.

\bibitem[Zhou et~al.(2023)Zhou, Li, Jiang, Wang, Zhou, Zhang, and Zhao]{dataset:pad}
Qiang Zhou, Weize Li, Lihan Jiang, Guoliang Wang, Guyue Zhou, Shanghang Zhang, and Hao Zhao.
\newblock {PAD:} {A} dataset and benchmark for pose-agnostic anomaly detection.
\newblock In \emph{Advances in Neural Information Processing Systems 36: Annual Conference on Neural Information Processing Systems 2023, NeurIPS 2023, New Orleans, LA, USA, December 10 - 16, 2023}, 2023.

\bibitem[Zhou et~al.(2024{\natexlab{b}})Zhou, Pang, Tian, He, and Chen]{AnomalyClip}
Qihang Zhou, Guansong Pang, Yu Tian, Shibo He, and Jiming Chen.
\newblock Anomalyclip: Object-agnostic prompt learning for zero-shot anomaly detection.
\newblock In \emph{The Twelfth International Conference on Learning Representations, {ICLR} 2024, Vienna, Austria, May 7-11, 2024}. OpenReview.net, 2024{\natexlab{b}}.

\bibitem[Zou et~al.(2022)Zou, Jeong, Pemula, Zhang, and Dabeer]{dataset:visa}
Yang Zou, Jongheon Jeong, Latha Pemula, Dongqing Zhang, and Onkar Dabeer.
\newblock Spot-the-difference self-supervised pre-training for anomaly detection and segmentation.
\newblock In \emph{European Conference on Computer Vision}, pages 392--408. Springer, 2022.

\end{thebibliography}
}


\clearpage

\setcounter{page}{1}
\maketitlesupplementary

\section{Comprehensive Detection Results}

We provide the full results for detection of all baselines on all classes of Real-IAD in~\cref{tab:big_table_samplewise} and~\cref{tab:big_table_imagewise}.
\vspace{2cm}

\begin{strip} 
\captionsetup{type=table}
\centering
\resizebox{\linewidth}{!}{

\begin{tabular}{l||c|c|c|c|c|c|c|c|c}
\multirow{2}{*}{Classes}    & PaDiM & Cflow & SoftPatch & DeSTSeg & RD   & UniAD & PatchCore    & SimpleNet &   \method          \\
 & \cite{padim}  & \cite{cflow} & \cite{softpatch} & \cite{destseg} & \cite{rd} & \cite{uniad} & \cite{patchcore} & \cite{simplenet} & (ours) \\\hline\hline
Audiojack         & \uline{92.2}          & 82.0         & 91.0         & 95.3                  & 81.9 & 91.2 & 89.3                  & 91.2                  & \textbf{96.56} $\pm$ 0.12 \\
Bottle Cap        & 98.1                  & 98.8         & 99.1         & 92.4                  & 93.7 & 97.3 & \textbf{\uline{99.4}} & \textbf{\uline{99.4}} & 98.00          $\pm$ 0.14 \\
Button Battery    & 88.7                  & 96.3         & 91.9         & 93.3                  & 83.3 & 87.5 & 90.6                  & \uline{95.8}          & \textbf{96.57} $\pm$ 0.16 \\
End Cap           & 76.1                  & 75.1         & \uline{92.3} & 82.3                  & 68.1 & 89.4 & 91.9                  & \textbf{94.2}         & 90.89          $\pm$ 0.09 \\
Eraser            & \textbf{96.5}         & 87.2         & 96.1         & 91.9                  & 82.9 & 91.2 & \uline{95.6}          & 94.7                  & 92.81          $\pm$ 0.14 \\
Fire Hood         & \textbf{\uline{96.9}} & 87           & 87.8         & \textbf{\uline{96.9}} & 81.4 & 83.0 & 89.3                  & 95.6                  & 95.45          $\pm$ 0.10 \\
Mint              & 69.1                  & 79.1         & 82.1         & 77.7                  & 67.7 & 73.0 & 85.7                  & \uline{86.8}          & \textbf{87.84} $\pm$ 0.03 \\
Mounts            & 98.4                  & 98.2         & 99.3         & 99.1                  & 92.5 & 97.0 & \textbf{99.7}         & 99.4                  & \uline{99.46}  $\pm$ 0.02 \\
PCB               & 88.4                  & 83.1         & 90.3         & 83.6                  & 79.3 & 83.2 & \uline{93}            & 90.7                  & \textbf{93.91} $\pm$ 0.11 \\
Phone Battery     & 91.7                  & 91.2         & 91.5         & \uline{98.2}          & 89.4 & 93.6 & 95.1                  & 94.7                  & \textbf{98.68} $\pm$ 0.06 \\
Plastic Nut       & \uline{98.2}          & 88.6         & 95.7         & 94.4                  & 72.8 & 87.1 & 97.8                  & 95.7                  & \textbf{98.96} $\pm$ 0.10 \\
Plastic Plug      & 87.4                  & 90.0         & 92.5         & \uline{95.6}          & 89.3 & 78.0 & \textbf{95.7}         & 94.4                  & 93.85          $\pm$ 0.07 \\
Porcelain Doll    & 93.8                  & \uline{95.1} & 94.7         & 94.6                  & 89.6 & 92.8 & 96.1                  & \textbf{96.2}         & 94.37          $\pm$ 0.42 \\
Regulator         & \textbf{96.5}         & 85.1         & 82.9         & 93.0                  & 92.5 & 55.5 & 86.0                  & 92.0                  & \uline{94.22}  $\pm$ 0.22 \\
Rolled Strip Base & 98.6                  & 98.8         & 99.7         & 98.9                  & 80.3 & 99.3 & \textbf{99.7}         & \uline{99.6}          & 99.57          $\pm$ 0.09 \\
SIM Card Set      & 94.2                  & 95.6         & 98.4         & 98.3                  & 89.9 & 94.0 & \textbf{99.3}         & \uline{99.2}          & 98.79          $\pm$ 0.04 \\
Switch            & 82.1                  & 92.9         & 97.8         & 96.6                  & 87.3 & 95.3 & 94.6                  & \uline{98.8}          & \textbf{99.27} $\pm$ 0.03 \\
Tape              & 99.8                  & 98.5         & 99.7         & 99.1                  & 89.5 & 99.1 & \uline{99.9}          & \textbf{100}          & 97.12          $\pm$ 0.12 \\
Terminal Block    & 96.9                  & 92.2         & \uline{98.2} & 96.1                  & 89.8 & 93.8 & 97.5                  & 97.7                  & \textbf{99.09} $\pm$ 0.02 \\
Toothbrush        & 91.7                  & 91.9         & 92.9         & \textbf{97.9}         & 86.7 & 95.0 & 94.7                  & 95.3                  & \uline{96.88}  $\pm$ 0.13 \\
Toy               & 91.4                  & 78.8         & 91.3         & \textbf{96.5}         & 75.0 & 77.2 & 92.8                  & 92.9                  & \uline{96.46}  $\pm$ 0.19 \\
Toy Brick         & 84.3                  & 82.9         & 78.2         & \uline{87}            & 72.5 & 78.3 & 82.6                  & 85.7                  & \textbf{91.73} $\pm$ 0.14 \\
Transistor1       & 90.3                  & 96.6         & 99.3         & 99.0                  & 94.7 & 99.3 & \textbf{99.8}         & \uline{99.7}          & 98.55          $\pm$ 0.07 \\
U Block           & 98.3                  & 96.7         & 98.3         & \uline{98.5}          & 86.9 & 96.3 & \textbf{98.8}         & \uline{98.5}          & 96.57          $\pm$ 0.10 \\
USB               & 77.0                  & 86.1         & 93.8         & 93.3                  & 89.4 & 83.1 & \uline{93.9}          & \uline{93.9}          & \textbf{96.83} $\pm$ 0.04 \\
USB Adaptor       & 93.2                  & 86.8         & 91.9         & \textbf{93.6}         & 65.3 & 85.1 & 90.6                  & 93.0                  & \uline{93.39}  $\pm$ 0.18 \\
Vcpill            & 94.7                  & 87.8         & 93.7         & 96.4                  & 87.2 & 89.4 & \uline{96.5}          & \textbf{97.5}         & 95.90          $\pm$ 0.11 \\
Wooden Beads      & 91.1                  & 89.3         & 90.9         & 91.9                  & 85   & 82.5 & 91.4                  & \uline{92.9}          & \textbf{94.72} $\pm$ 0.16 \\
Woodstick         & 81.8                  & 83.9         & 73.9         & \textbf{90.2}         & 71.9 & 76.0 & 74.5                  & 81.5                  & \uline{89.15}  $\pm$ 0.19 \\
Zipper            & 99.3                  & 97.6         & 99.6         & 99.7                  & 96.1 & 98.8 & \textbf{100}          & 99.7                  & \uline{99.92}  $\pm$ 0.00 \\\hline
Average           & 91.2                  & 89.8         & 92.8         & 94.0                  & 83.7 & 88.1 & 93.7                  & \uline{94.9}          & \textbf{95.85} $\pm$ 0.02 

\end{tabular}
}
\captionof{table}{Performance in sample-wise anomaly detection on all classes of Real-IAD~\cite{realiad}. Here, the anomaly scores for all images belonging to the same object are aggregated into one value. Best performances are \textbf{bold}, with the runner-up \underline{underlined}. We execute our method for $n=5$ runs and report mean $\pm$ standard deviation.}
\label{tab:big_table_samplewise}
\end{strip}

\begin{table*}
\centering
\resizebox{\linewidth}{!}{
\begin{tabular}{l||c|c|c|c|c|c|c|c|c}
\multirow{2}{*}{Classes}    & PaDiM & Cflow & SoftPatch & DeSTSeg & RD   & UniAD & PatchCore    & SimpleNet &   \method          \\
 & \cite{padim}  & \cite{cflow} & \cite{softpatch} & \cite{destseg} & \cite{rd} & \cite{uniad} & \cite{patchcore} & \cite{simplenet} & (ours) \\\hline\hline
Audiojack         & 66.6 & 74.3 & \uline{88.5}  & 81.8          & 82.4          & 82.8 & 86.3          & 87.4          & \textbf{90.69} $\pm$ 0.06 \\
Bottle Cap        & 87.7 & 91.3 & \textbf{95.9} & 87.1          & 89.2          & 89.8 & 94.3          & 91.6          & \uline{94.62}  $\pm$ 0.14 \\
Button Battery    & 84.8 & 85.2 & 88.5          & \textbf{91.2} & 87.0          & 79   & 87.3          & 88.8          & \uline{89.37}  $\pm$ 0.16 \\
End Cap           & 73.7 & 66.4 & \textbf{85.8} & 80.3          & 79.0          & 80.4 & 84.1          & 83.4          & \uline{84.77}  $\pm$ 0.10 \\
Eraser            & 86.7 & 88.1 & \textbf{93.5} & 88.2          & 89.2          & 89.6 & \uline{93.4}  & 91.2          & 92.08          $\pm$ 0.11 \\
Fire Hood         & 77.3 & 80.5 & \uline{84.3}  & 78.9          & \uline{84.3}  & 79.5 & 84.1          & 81.8          & \textbf{93.35} $\pm$ 0.13 \\
Mint              & 66.9 & 70.7 & 74.5          & 70.5          & 71.6          & 67.6 & 76.2          & \uline{77.2}  & \textbf{79.38} $\pm$ 0.11 \\
Mounts            & 82.5 & 85.3 & 85.9          & 85.1          & 85.7          & 87.2 & \uline{88}    & \textbf{88.2} & 84.26          $\pm$ 0.11 \\
PCB               & 75.0 & 77.3 & 90.8          & 79.6          & 89.5          & 80.5 & \textbf{92.4} & 87.1          & \uline{91.79}  $\pm$ 0.09 \\
Phone Battery     & 81.9 & 84.4 & 90.2          & 86.2          & \uline{90.8}  & 83.4 & \textbf{91.6} & 88.9          & 90.77          $\pm$ 0.35 \\
Plastic Nut       & 73.8 & 79.8 & 89.3          & \uline{90.2}  & 85.0          & 79.3 & \textbf{90.8} & 89.3          & 88.04          $\pm$ 0.13 \\
Plastic Plug      & 80.0 & 83.9 & 88.7          & 86.0          & \textbf{90.5} & 78.2 & \uline{89.7}  & 87.1          & 88.45          $\pm$ 0.11 \\
Porcelain Doll    & 74.3 & 76.0 & 86.1          & 83.7          & \uline{87.8}  & 84.1 & \textbf{88.2} & 86.1          & 86.62          $\pm$ 0.23 \\
Regulator         & 76.5 & 62.9 & 82.1          & \textbf{89.8} & 87.3          & 51.8 & 81.9          & 82.2          & \uline{87.69}  $\pm$ 0.18 \\
Rolled Strip Base & 97.4 & 97.1 & \uline{99.1}  & 98.3          & 94.3          & 98.6 & 98.9          & \textbf{99.4} & 98.15          $\pm$ 0.12 \\
SIM Card Set      & 91.8 & 94.4 & 95.8          & 91.8          & 93.6          & 91.1 & \textbf{97.1} & \uline{96.1}  & 90.65          $\pm$ 0.16 \\
Switch            & 80.8 & 83.9 & 94.7          & 92.5          & 87.2          & 85.7 & 89.4          & \uline{94.8}  & \textbf{96.03} $\pm$ 0.12 \\
Tape              & 93.7 & 96.0 & \uline{97.9}  & 96.1          & 93.1          & 97.2 & \textbf{98.2} & 96.9          & 94.32          $\pm$ 0.11 \\
Terminal Block    & 86.7 & 85.1 & \uline{96.2}  & 91.4          & 94.5          & 85.9 & 94.1          & 93.4          & \textbf{96.41} $\pm$ 0.19 \\
Toothbrush        & 74.4 & 79.9 & \textbf{89.3} & 87.5          & 83.4          & 82.7 & \uline{88.0}  & 86.1          & 85.21          $\pm$ 0.17 \\
Toy               & 82.5 & 67.8 & 86.7          & 82.1          & 81.5          & 67.3 & \uline{87.3}  & 82.2          & \textbf{88.31} $\pm$ 0.10 \\
Toy Brick         & 71.0 & 80.7 & 79.8          & 76.8          & 73.9          & 77.9 & \uline{81.4}  & 80.8          & \textbf{87.35} $\pm$ 0.01 \\
Transistor1       & 86.5 & 92.8 & 97.2          & 95.6          & 95.8          & 94.1 & \textbf{97.8} & \uline{97.3}  & 94.86          $\pm$ 0.14 \\
U Block           & 84.1 & 90.5 & 91.6          & 90.1          & \uline{91.7}  & 88.8 & \textbf{91.9} & 90.7          & 90.45          $\pm$ 0.19 \\
USB               & 69.7 & 78.1 & 91.2          & 90.3          & \uline{92.1}  & 80.9 & 90.7          & 90.0          & \textbf{93.37} $\pm$ 0.03 \\
USB Adaptor       & 81.9 & 75.2 & \uline{86}    & 73.1          & 71.9          & 77.3 & 83.8          & 82.7          & \textbf{88.07} $\pm$ 0.13 \\
Vcpill            & 67.4 & 85.8 & 89.6          & 90.0          & 90.2          & 88.6 & \textbf{91.4} & \uline{91.1}  & 89.50          $\pm$ 0.19 \\
Wooden Beads      & 82.4 & 87.4 & \textbf{89.2} & 86.2          & 87.4          & 80.7 & \uline{88.6}  & 85.7          & 87.28          $\pm$ 0.05 \\
Woodstick         & 79.9 & 78.9 & 76.0          & \textbf{89}   & 84.2          & 78.9 & 77.2          & 77.9          & \uline{88.04}  $\pm$ 0.17 \\
Zipper            & 90.4 & 93.8 & 97.9          & 98.4          & 97.8          & 98.2 & 98.1          & \uline{98.2}  & \textbf{98.23} $\pm$ 0.08 \\\hline
Average           & 80.3 & 82.5 & \uline{89.4}  & 86.9          & 87.1          & 82.9 & \uline{89.4}  & 88.5          & \textbf{90.27} $\pm$ 0.02 
\end{tabular}
}
\caption{Performance in image-wise anomaly detection on all classes of Real-IAD~\cite{realiad}. Here, the anomaly scores for all images belonging to the same object are aggregated into one value. Best performances are \textbf{bold}, with the runner-up \underline{underlined}. We execute our method for $n=5$ runs and report mean $\pm$ standard deviation.}
\label{tab:big_table_imagewise}
\end{table*}

\begin{table*}[ht]
\centering
\resizebox{\textwidth}{!}{
\begin{tabular}{l||c|c|c|c|c|c|c|c|c}
\multirow{2}{*}{Metrics}    & PaDiM & Cflow & SoftPatch & DeSTSeg & RD   & UniAD & PatchCore    & SimpleNet &   \method          \\
 & \cite{padim}  & \cite{cflow} & \cite{softpatch} & \cite{destseg} & \cite{rd} & \cite{uniad} & \cite{patchcore} & \cite{simplenet} & (ours) \\\hline\hline

AUPRO       & 81.8  & 90.5  & 90.8      & 81.5    & \textbf{93.8} & 86.1  & \underline{91.5}         & 84.6            & 87.91   $\pm$ 0.01        \\
Pixel-AUROC  & -     & -     & -         & -       & -    & -     & -            &   - &         96.47 $\pm$ 0.00 \\
\end{tabular}
}
\caption{\textbf{Average} segmentation performance of various methods across all classes in Real-IAD~\cite{realiad}. Pixel-wise AUROC has not been reported for the baselines. Best performances are \textbf{bold}, with the runner-up \underline{underlined}. We execute our method for $n=5$ runs and report mean $\pm$ standard deviation.}
\label{tab:appendix_segmentation}
\end{table*}

\end{document}